\documentclass{ieeeaccess}
\usepackage{cite}
\usepackage{amsmath,amssymb,amsfonts}
\usepackage{algorithmic}
\usepackage{graphicx}
\usepackage{textcomp}
\usepackage[linesnumbered,ruled,vlined]{algorithm2e}
\usepackage{caption}
\usepackage[caption=false,font=footnotesize]{subfig}
\makeatletter
\newcommand*{\rom}[1]{\expandafter\@slowromancap\romannumeral #1@}
\makeatother

\def\BibTeX{{\rm B\kern-.05em{\sc i\kern-.025em b}\kern-.08em
    T\kern-.1667em\lower.7ex\hbox{E}\kern-.125emX}}

\begin{document}
\history{Date of publication xxxx 00, 0000, date of current version xxxx 00, 0000.}
\doi{10.1109/ACCESS.2017.DOI}

\title{Joint-SRVDNet: Joint Super Resolution and Vehicle Detection Network}
\author{\uppercase{Moktari Mostofa}\authorrefmark{1}, \IEEEmembership{Student Member, IEEE},
\uppercase{Syeda Nyma Ferdous}\authorrefmark{1}, 
\uppercase{Benjamin S. Riggan}\authorrefmark{2},
\IEEEmembership{Member, IEEE}, and
\uppercase{Nasser M. Nasrabadi}\authorrefmark{1},
\IEEEmembership{Fellow, IEEE}}
\address[1]{Lane Department of Computer Science and Electrical  Engineering, West Virginia University, Morgantown, WV 26506, USA}
\address[2]{Department of Electrical and Computer Engineering, University of Nebraska-Lincoln, Lincoln, NE 68588, USA}
\markboth
{M. Mostofa \headeretal: Joint-SRVDNet: Joint Super Resolutionand Vehicle Detection Network}
{M. Mostofa \headeretal: Joint-SRVDNet: Joint Super Resolutionand Vehicle Detection Network}

\corresp{Corresponding author: Moktari Mostofa (e-mail: mm0251@mix.wvu.edu)}

\begin{abstract}
In many domestic and military applications, aerial vehicle detection and super-resolution algorithms are frequently developed and applied independently. However, aerial vehicle detection on super-resolved images remains a challenging task due to the lack of discriminative information in the super-resolved images. To address this problem, we propose a Joint Super-Resolution and Vehicle Detection Network (Joint-SRVDNet) that tries to generate discriminative, high-resolution images of vehicles from low-resolution aerial images. First, aerial images are up-scaled by a factor of 4x using a Multi-scale Generative Adversarial Network (MsGAN), which has multiple intermediate outputs with increasing resolutions. Second, a detector is trained on super-resolved images that are upscaled by factor 4x using MsGAN architecture and finally, the detection loss is minimized jointly with the super-resolution loss to encourage the target detector to be sensitive to the subsequent super-resolution training. The network jointly learns hierarchical and discriminative features of targets and produces optimal super-resolution results. We perform both quantitative and qualitative evaluation of our proposed network on VEDAI, xView and DOTA datasets. The experimental results show that our proposed framework achieves better visual quality than the state-of-the-art methods for aerial  super-resolution with 4x up-scaling factor and improves the accuracy of aerial vehicle detection.
\end{abstract}

\begin{keywords}
Aerial images, Multi-scale Generative Adversarial Network (MsGAN), super-resolution, vehicle detection.
\end{keywords}

\titlepgskip=-15pt
\maketitle
\section{Introduction}


Real-time vehicle detection in aerial imagery has been an active research area in recent years \cite{cao2019affine, ferdous2019super, aspiras2019convolutional, yang2019vehicle}. Due to high altitudes in which aerial images are acquired, targets of interest (e.g., vehicles) contain fewer pixels than targets imaged at considerably lower elevations (e.g., building surveillance cameras, or traffic cameras), which significantly degrades detection performance. Moreover, complex background and computational constraints further hinder detection performance.
Single image super-resolution (SISR) techniques are commonly used to alleviate poor detection performance by generating a
high-resolution counterpart to the original low-resolution image. Recently, generative adversarial networks (GANs) \cite{goodfellow2014generative} have demonstrated the ability to synthesize high-quality images \cite{ledig2017photo, hoffman2017cycada} for many applications, including super-resolution. However, GANs have also been known to be somewhat unstable, frequently lacking discriminability in synthesized imagery. Therefore, we aim to produce and simultaneously train both discriminative and super-resolved images by using multi-task learning to combine correlated tasks such as super-resolution and object detection networks.

The inter-relationship between super-resolution techniques and object detection algorithms has been previously studied  to improve detection performance \cite{ferdous2019super, shermeyer2019effects, borel2019image}. However, none of them have tried to explore performance of super-resolution if the entire network is trained jointly. One might presume that the reason there are still misdetections and detection failures is because the super-resolution algorithm is not optimized for target detection task.

In this paper, we propose a deep neural network (DNN) framework to simultaneously generate super-resolved aerial images and locate vehicles in the super-resolved images. Our proposed framework is composed of (i) a Multi-scale Generative Adversarial Network (MsGAN) framework to create super resolved versions of the original images. This network preserves high-level features when mapping between low resolution to high resolution domains, and (ii) locate vehicles using one of the variants of YOLO \cite{redmon2016you} introduced in \cite{redmon2018yolov3} as YOLOv3 object detector. We jointly train the entire network at each iteration such that target regions in the super-resolved images become contextually more distinctive from the background. We refer to our proposed algorithm in this paper as the Joint Super-Resolved Vehicle Detection Network (Joint-SRVDNet). Our proposed framework has been evaluated on several extensively used aerial datasets. We train the model on VEDAI, xView and DOTA datasets to evaluate both qualitative and quantitative performances. Moreover, our network shows promising performances compared to a set of state-of-the-art methods. In summary, the key contributions of this paper are as follows:

\begin{itemize}

\item In this paper, we propose an end-to-end jointly trainable deep neural network what we named Joint-SRVDNet, which offers a multi-tasking paradigm by handling both super-resolution and vehicle detection for aerial and satellite imagery. To the best of our knowledge, our proposed Joint-SRVDNet is the first multi-task model that leverages complementary information of the two tasks to jointly learn Super-Resolution (SR) and vehicle detection in aerial images. Such a novel framework allows for improved super-resolution reconstructions and more accurate vehicle detection in aerial imagery. 

\item An MsGAN architecture is proposed for the first time for aerial and satellite image super-resolution, which ensures  progressive  learning  of  the statistical  distributions  of images  at  multi-scale  and significantly improves the performance of SR reconstruction by producing discriminative and high-quality super-resolved images.

\item The proposed MsGAN architecture for super-resolution has potential  contributions  to  vehicle  detection  in  low-resolution aerial and satellite images.

\item We show remarkable improvements for both super-resolution and vehicle detection for low-resolution aerial imagery with comparable performance to the existing state-of-the-art methods when evaluated on the corresponding high-resolution aerial images. 

\end{itemize}

The rest of this paper is organized in the following manner. Section II reviews related super-resolution and detection algorithms. It also describes challenges when applied to aerial imagery. We give details of our proposed method in section III. Besides, we also discuss the training loss functions of our network in section IV. Section V presents the datasets and experimental details of our work. Section VI shows comparative results and explains the performance. Finally, we provide a conclusion and state some limitations of our algorithm in section VII. 

\section{RELATED WORK}
\subsection{Deep Learning Based Single Image Super-Resolution}
Single Image Super-Resolution (SISR) techniques have been studied extensively in the field of computer vision. Recently, Convolutional Neural Network (CNN) architectures have been widely used in image SR algorithms since they can extract representative features that are useful in recovering high-frequency details in super-resolved images. A three-layer CNN was first proposed by Dong et al. \cite{dong2015image} and referred as SRCNN to learn a mapping between Low-Resolution (LR) and High-Resolution (HR) image pairs, which was later modified in VDSR \cite{kim2016accurate} and DRCN \cite{kim2016deeply}. In VDSR \cite{kim2016accurate}, Kim et al. implemented an efficient SSIR method, where they showed that increasing the network depth trained by adjustable gradient clipping resulted in a significant improvement in visual quality of super-resolved images. In DRCN \cite{kim2016deeply}, they increased recursion depth by adding more weight layers with skip connection to improve the performance of SRCNN. 
However, all these methods apply interpolation to the LR inputs, which significantly loses some useful information and thereby yields poor results with increased computational cost. Since then these super-resolution architectures have been frequently modified by developing CNN-based architectures like Residual Networks (ResNet) \cite{he2016deep}, Recurrent Neural Networks (RNNs) \cite{eigen2013understanding, liang2015recurrent, socher2012convolutional} to extract features from the original LR inputs. 

Recently, GANs \cite{goodfellow2014generative} have replaced these SR algorithms. Ledig et al. \cite{ledig2017photo} introduce ResNet as the base architecture for image super-resolution and utilize the idea of GAN to reconstruct fine texture details in the super-resolved images. GAN architectures have successfully attained superior performances in many applications of computer vision, such as style transfer, image reconstruction and image SR. SRGAN \cite{ledig2017photo} is the first attempt which utilizes GAN to produce photo-realistic natural looking images close to the original high resolution images. They formulate a loss function which is a combination of a perceptual similarity loss \cite{bruna2015super, gatys2015texture, johnson2016perceptual} in addition to an adversarial loss \cite{goodfellow2014generative} so that the network learns to preserve content of images during SR training. Although SRGAN has shown remarkable performances, still it finds difficulty in generating high-resolution (e.g., $256\times256$) images due to  training instability and mode collapse. During upscaling the LR images to the desired HR counterparts, GAN suffers from the training instability due to low chance of sharing hyper-parameters between image distribution and model distribution in a high-dimensional space. To stabilize the training process, Zhang et al. proposed  StackGAN \cite{zhang2017stackgan}. The motivation came from the observation that image distributions are related at multiple scales. StackGAN outperforms significantly other state-of-the-art methods in reconstructing real looking super-resolved images. In StackGAN, they used multiple-generators along with discriminators at each scale to  share most of their parameters across the whole network. This structure pushes the resulting solutions towards the original image distributions. For our work, we incorporate the idea of using  multiple discriminators at each different scale in addition to the work of Ledig et al. \cite{ledig2017photo} where the authors use a perceptual loss function with Mean Squared Error (MSE) loss to generate more realistic SR images. Our network can be viewed as  multi-scale GAN architecture since we are using only one generator instead of multiple generators like StackGAN and stack discriminators at each intermediate outputs to improve the learning of image distributions at multiple scales. As shown in Fig. 2, discriminators at intermediate outputs sequentially help generator produce real-looking super-resolved images to the desired size. The prime goal is to approximate highly related image distributions at different scales. So, stacking multiple discriminators helps the network accomplish this goal by continuously giving feedback from image distributions at one scale to another. 

\subsection{Deep Learning Based Vehicle Detection Architectures}
Vehicle detection recently has become a prominent research area with applications in civilian and military surveillance, traffic monitoring and planning transportation systems. In \cite{zhao2003car}, the authors proposed a method which utilized Bayesian network to integrate the important features for car detection. Choi et al. \cite{choi2009vehicle} applied the Mean-shift algorithm to extract car like shape for detecting cars in satellite images. In the work of \cite{cheng2012vehicle}, they trained a Dynamic Bayesian Network (DBN) to preserve region level features. 

Carlet and Abayowa \cite{DBLP:journals/corr/abs-1709-08666} proposed a modified YOLOv2 \cite{sang2018improved} for locating vehicles in aerial imagery. A modified faster R-CNN was applied in the work of Terrail et al. \cite{terrail2018faster} that showed promising performances in aerial vehicle detection. In \cite{soleimani2018convolutional}, Soleimani et al. proposed a text-guided detection scheme using both visual and textual features for detection. 
Yang et al. \cite{yang2018vehicle} applied skip connection in their framework to merge lower and higher level features and utilized a focal loss function for vehicle detection. For multi-oriented vehicle detection, Li et al. \cite{deng2018r3net} designed a rotatable region proposal network which learned the orientation of vehicles while 
performing classification on aerial images and videos.

Vehicle detection in overhead imagery remains a challenging issue due to the low resolution of vehicles. To alleviate this shortcoming, researchers have focused on super-resolution techniques. An overview of detection performance on super-resolved images is reported in \cite{shermeyer2019effects} considering multiple-resolutions. In this paper, we propose a joint training approach which learns to extract discriminative features from low-resolution images such that it can produce super-resolved images that are as visually similar to the corresponding high-resolution images as possible. 

\subsection{Joint Training of Super-Resolution and Detection}
Improving object detection performance guided by learning based super-resolution has been a recent research focus. In \cite{shermeyer2019effects}, the impact of super-resolution on object detection has been extensively studied. Haris et al. \cite{haris2018task} adopt a task-driven super resolution approach employing a novel compound loss based end-to-end training that enhances the image quality leading to a better recognition. Cansizoglu et al. \cite{ataer2019verification} design an identity preserving face super-resolution framework and achieve outstanding  performance for face verification in real time. In this work, the authors propose to  use a two-stage loss minimization technique rather than end-to-end training. They hypothesize that end-to-end training involves higher computational complexity respect to limited data samples. On the other hand, another study in \cite{wu2016deep} propose a deep model that jointly optimizes face hallucination and verification loss for low resolution face identification. In this study, face hallucination loss is measured in terms of pixel difference between the ground truth HR images and network-generated images and verification loss is estimated by the classification error and intra-class distance. Most of the recent works focus on verification, which is easier from the detection task. For example, verification confirms identity whereas detection involves recognition of desired object (e.g., human face, vehicle, etc.). Again, during verification, the probe face has already been detected, but detection has to minimize different constraints before detecting the target object. 

Pang et al. \cite{pang2019jcs} introduce JCS-Net that combines classification and super-resolution task as one for small-scale pedestrian detection. However, these algorithms do not deal with vehicle detection and super-resolution for aerial imagery that deals with  more fundamental challenges. For instance, the average height of pedestrians in the benchmark datasets (e.g., Caltech \cite{dollar2009pedestrian}, KITTI \cite{geiger2012we}) ranges from 60 pixels tall to 430 pixels tall, whereas the average resolution for aerial vehicles is $10 \times 15$ pixels in the publicly available benchmark datasets (e.g., VEDAI \cite{razakarivony2016vehicle}, xVIEW \cite{xia2018dota}, DOTA \cite{lam2018xview}), which yields poor detection results. 

These reviews strongly suggest to use super-resolution technique for developing a robust detection system, which helps to recover detailed information in the low-resolution space. In this paper, we try to investigate the relationship between super-resolution and vehicle detection by proposing a joint training approach so that they can be benefited from each other. We propose to integrate both super-resolution and detection network together. Usually, the super-resolution technique recovers useful detailed information in the low-resolution image, but here it focuses especially on the target regions as detector loss is integrated to SR training. The network gradually learns the input image distributions in the high-resolution space and produce super-resolved version of low-resolution image with distinctive properties of target objects, which also helps detector to achieve better results.

\begin{figure*}
\centering
\includegraphics[width=15cm]{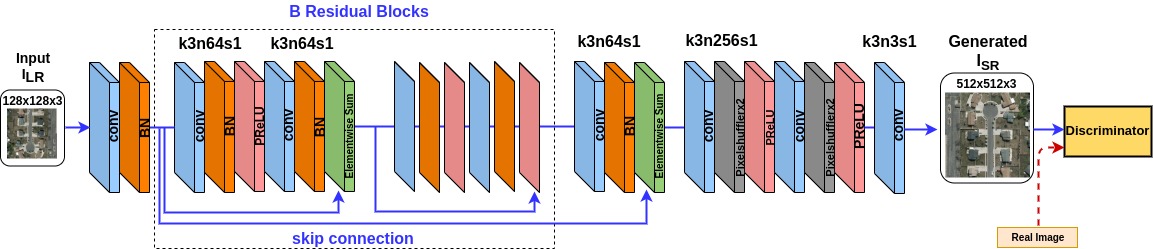}
\caption{Architecture of SRGAN with corresponding kernel size (k), number of feature maps (n) and stride (s) indicated for each convolutional layer.}
\end{figure*}

\begin{figure*}
\centering
\includegraphics[width=15cm]{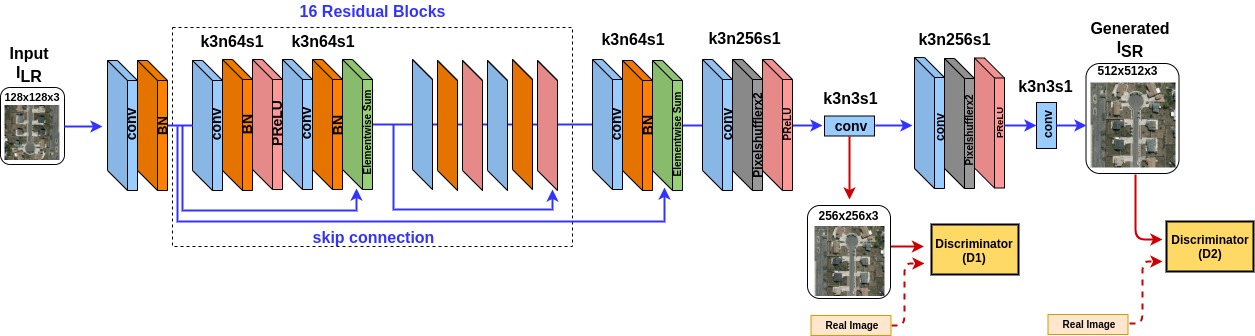}
\caption{Architecture of Multi-scale SRGAN (MsSRGAN) with corresponding kernel size (k), number of feature maps (n) and stride (s) indicated for each convolutional layer.}
\end{figure*}
  
\section{Proposed Framework}
In this section, we describe our proposed framework in detail. The proposed framework is an end-to-end network that generates super-resolved aerial images using an MsGAN architecture and jointly optimized YOLOv3 detector to perform vehicle detection in aerial super-resolved imagery.

\subsection{Generative Adversarial Networks (GANs)}

GANs are a special type of generative models which have shown remarkable performances in representation learning and synthesized image generation. They have been widely used in image super-resolution (first applied by Ledig et al. in \cite{ledig2017photo}), image synthesis and image translation using conditional GANs (cGANs) \cite{isola2017image} and cyclic GANs (cycleGANs) \cite{hoffman2017cycada}. Their goal is to learn statistical distribution of the training data to train a mapping $G:x\rightarrow y$ such that image distribution from $G(x)$ is indistinguishable from image distribution of target $y$. Typically, the generator  $G$ is a differentiable function which is trained to learn the distribution $p_{data}$ over data $y$. To do so, it takes input from the distribution $p_{x}(x)$ and maps it to the target data space as $G(x; \theta_{g})$ where $\theta_{g}$ defines the parameters of the generator model. In addition, the discriminator $D$ acts like a classifier which is trained to return probability distributions $D(y)$ and $D(G(x))$ for both training examples from the distribution $p_{data}(y)$ and samples from $G(x)$, respectively. Basically, $D$ is trained to maximize the probability of assigning the correct label to both training examples and samples from $G$. Simultaneously $G$ is trained to minimize $log(1 - D(G(x))$). In other words, $D$ and $G$ play the following two-player minimax game with the adversarial loss $l_{GAN}(G, D)$:
\begin{equation}
\begin{split}
\min_{G}\max_{D} l_{GAN}(G,D) &= \min_{G}\max_{D}[E_{y\sim p_{data}}[log D(y)]\\&\quad+E_{x\sim p_x}[log(1-D(G(x)))]].
\end{split}
\end{equation} However, it is very difficult to achieve the desired output by training the network only with adversarial loss. Adding a $\textit{l}_{L1}$ reconstruction loss in addition to adversarial loss may result in high quality super-resolved images. Thus, the final objective function consists of two loss function as follows:

\begin{equation}
 G^* = arg  \min_{G} \max_{D} l_{GAN}(G,D) + \lambda l_{L1}(G),
\end{equation}where $l_{L1}(G)=\frac{1}{N}\sum\limits_{i=1}^N{\vert\vert y_{i} - G(x_{i}) \vert\vert}_1$, $N$ defines the number of samples in the training set  and  $\lambda$  is a weighting factor.

\subsection {Multi-scale GAN Architecture for Image Super-Resolution}
One of the objectives of our work is to estimate a high resolution version with distinctive features of its low resolution input aerial image. The network is trained to learn a generating function $G$ that aims to output photo-realistic images (according to a large distribution of images). Our basic deep generator network is illustrated in Fig. 1 which consists of B(=16) serially connected residual blocks with identical layout. Each residual block uses two convolution layers of 3x3 kernel and 64 feature maps followed by batch-normalization layers \cite{ioffe2015batch} and ParametricReLU \cite{he2015delving} as the activation function. To increase the resolution of the input image, we employ two sub-pixel convolutional layers \cite{shi2016real} in our generator network. 

Although this architecture achieved promising results in recovering high-frequency information from low-resolution images; it cannot handle varying condition (sharpness, atmospheric turbulance, motion blur, etc.). Usually, the estimated super-resolved images suffer from image blurriness and shape distortions. Moreover, some details which are vital for producing natural looking images are missing in the super-resolved images. 

\begin{figure*}
\centering
      \includegraphics[width=17cm]{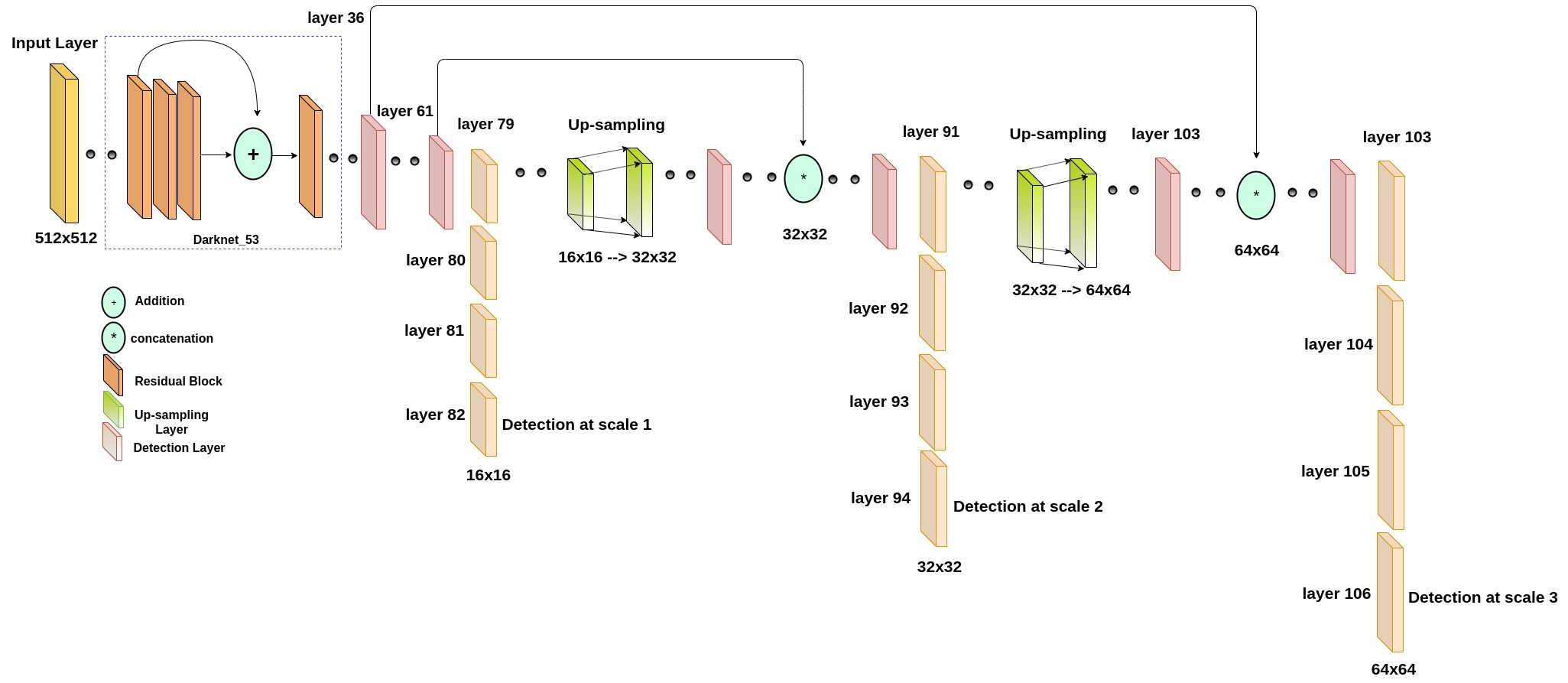}
  \caption{YOLOv3 architecture for vehicle detection at three scales showing residual block, upsampling layer as feature extractor.}
\end{figure*}

One application of aerial image super-resolution is vehicle detection, which requires enough visual detail to distinguish vehicles from background (e.g., roads, buildings, trees, etc.) in super-resolved images. Our  previous detection results \cite{ferdous2019super} showed that this network is not able to produce a high-detection performance while performing on super-resolved images generated by the classical SRGAN. We follow the framework of Kazemi et al. \cite{kazemi2019identity} and Wang et al. \cite{wang2018aerial} to build a progressive generator that learns to reconstruct a multi-stage network through a series of multi-scale image reconstructions. We train our generator model to produce multiple outputs at different resolutions as shown in Fig. 2. The main idea is to encourage the network to learn the image distribution at different scales. We enforce constraints on our network at two different image resolutions 256x256 and 512x512. When the network generates images of size 256x256, the first discriminator, $D1$ is pushing the generator to learn the probability distribution at that scale. Simultaneously, the second discriminator, $D2$ is contributing to help the generator to learn the distribution of the training images of size 512x512.

Gradually, the network learns to remove blurriness and recover missing object parts as it is trained at multi scales. Following this approach, it assures information transfer between images of different scales and generate more high-quality images. 

We follow similar network structure for both discriminators $D1$ and $D2$. We adapt the architectural guidelines from Radford et al. \cite{radford2015unsupervised} to design our discriminator. We utilize a LeakyReLU activation $(\alpha = 0.2)$  and avoid max-pooling to ignore feature size reduction. Our discriminator has eleven convolutional layers, which use 4x4 filter kernels.  Network employs strided convolutions to decrease image resolution while increasing the feature map size. At the end of the network, one dense layer and a final sigmoid activation function is added to obtain a probability for sample classification. 
\subsection {Aerial Vehicle Detection}
Our goal is to perform vehicle detection on several aerial datasets. The datasets contain aerial vehicles of different sizes which require strong detection algorithm to extract contextual and semantic information of those target objects. In our research work,  we use YOLOv3 of the state-of-the-art object detection algorithms to perform vehicle detection in real-time. 
\subsubsection{Architecture Details}
The architecture of YOLOv3 shown in Fig. 3 is based on the idea of residual network which employs Darknet-53 convolutional network for feature extraction.  To retrieve fine-grained information, it concatenates deeper layers with the earlier layers through up-sampling.  YOLOv3 takes an image and divides it into $M \times M$ ($16 \times 16$, $32 \times 32$ and $64 \times 64$ as in Fig. 3) grids. Then it applies classification and localization at each grid size. The grid cell is responsible for detecting object, if the center of the ground truth object falls within a grid cell. For each grid cell, a number of bounding boxes with their confidence scores and their associated class probabilities are generated using a fully convolutional network architecture.
YOLOv3 performs multi-scale prediction applying the feature pyramid network (FPN) \cite{lin2017feature} concept. It predicts objects at three different scales of 16, 32 and 64 for large, medium and small object detection. YOLOv3 uses 9 anchor boxes while predicting objects. Design of the anchor boxes greatly impacts the performance of the detector. We have used k-means clustering to generate these anchors for each database. The final number of detection results by YOLOv3 is $M\times M\times (B*(4+1+C))$. Here, $M\times M$ is the number of grid cells, $B$ is predicted number of bounding boxes in a cell, 4 denotes the four coordinates of the bounding boxes and 1 is for the objectness score, $C$ is the number of classes ( $C$=1:'vehicle' in our experiments). It uses multi-label classification. Softmax is replaced by a logistic regression to compute objectness score. Instead of using mean squared error in calculating the classification loss, it uses the binary cross-entropy loss for each label.  

\subsection{Our Proposed Joint Super-Resolution and Detection Network}
In  this  paper,  we  propose  an  end-to-end  multi-task model that jointly does super-resolution and vehicle detection in aerial imagery. Super-Resolution and vehicle detection for low-resolution aerial images have been considered as highly interrelated tasks. Usually, multi-task learning is adapted to address such highly correlated tasks as they can leverage significant information from each other. The vehicles in aerial scenes suffer from appearance ambiguity due to the low resolution characteristics of the images. In addition, it becomes challenging to deal with different sizes of vehicles with varying conditions such as blurry edges and lack of sharpness, etc. Moreover, the similarities between target vehicles and complex background make it even more difficult during detection. 

In our previous work \cite{ferdous2019super}, super-resolution and vehicle detection networks were developed independently to help each other. We notice that the information extracted from the low-resolution space is not maximized when only one task is performed without utilizing the advantages of the other task (e.g., detection is performed on super-resolved images generated from already trained SR module). In other words, if we apply super-resolution and vehicle detection successively, it does not benefit from multi-tasking. Therefore, our goal is to create a bridge between these highly interrelated tasks so that they can get the maximum benefit from the multi-task learning. Hence, we propose the Joint-SRVDNet to generate distinctive super-resolved images with high perceptual quality and simultaneously locate vehicles on these super-resolved images. We  have  developed  a  MsGAN super-resolution module that explicitly incorporates the structural information  (edges,  sharpness,  perceptual  features  defined by visual  deterministic properties  of  objects)  about targets into the  super-resolution reconstruction  process  as  well  as jointly learns both the super-resolution and object detection modules together as presented in Fig. 4. As shown in Fig. 4 super-resolution and detection modules are cascaded to execute the joint training in an end-to-end fashion. 

The joint loss optimization of our model is difficult to converge from scratch compared to the training of each module independently. Therefore, we first train super-resolution module given the paired high-resolution and corresponding low-resolution aerial training images. Then we train detection module with high resolution images to obtain network parameters for further training. Finally, we fine tune both modules together and integrate into one unified framework by optimizing (7) where super-resolution and detection losses are jointly trained together. Such a training scheme leads to a better convergence. Our proposed network optimizes a combination of four different losses : adversarial loss, pixel-wise mean square error (MSE), perceptual loss, and detection loss. The adversarial loss aims to help generator to create solutions that are close to real images by differentiating between real and generated aerial images. The widely used pixel-wise MSE estimates an overly smoothed solution as it only  measures pixel differences between super-resolved images and ground truth high resolution images. A perceptual loss using the pretrained VGG-19 network recovers photo-realistic textures, and a detection loss that aims for locating the target of interests with varying attributes such as lost edge details and structural features.

\begin{figure}
  \centering
      \includegraphics[width=9cm]{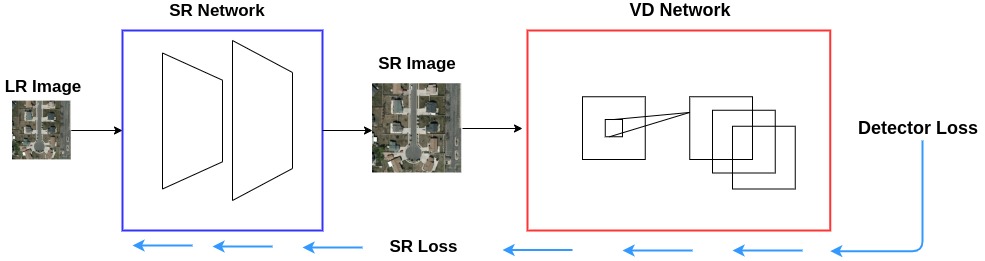}
       \caption{Architecture of our proposed model Joint-SRVDNet during the training process where the detector and super-resolution losses are back propagated to the generator.}
  
\end{figure}
\section{Loss Function}
We combine multiple loss terms to train our proposed joint network. The ultimate final loss function includes pixel-wise MSE loss, perceptual loss, adversarial loss and detection loss. 

\subsection{Pixel-Wise MSE Loss}

State-of-the-art image SR methods \cite{dong2015image, shi2016real} mostly rely on pixel-wise MSE loss to optimize the network. For the training images $I_{n}^{HR}$ with their corresponding low-resolution $I_{n}^{LR}$, \textit{n} = \textit{1,....,N},
we can calculate the MSE loss also referred to as the content loss $l_{cont}$ using the following equation:
\begin{equation}
L_{cont} = \frac{1}{N}\sum\limits_{n=1}^N\frac{1}{WH}\sum\limits_{x=1}^W \sum\limits_{y=1}^H ((I_{n}^{HR})_{x,y}-G(I_{n}^{LR})_{x,y})^2,
\end{equation}where W and H represent width and height of the image and $G(I_{n}^{LR})$ are the super-resolved images for \textit{N} training samples. 

Although MSE loss is the widely used optimization method for super-resolution which achieves high peak signal-to-noise ratios, the resulting estimates often lack fine texture details and are perceptually not convincing because of overly blurry results. In addition, MSE doesn't have ability to capture spatially varying high frequency information, as it is based on pixel-wise image differences.  
\subsection{Perceptual Loss}
Since optimizing the MSE loss is prone to overfitting when defined over the pixel-wise differences between estimated super-resolved images and ground truth high resolution images, Ledig et al. \cite{ledig2017photo} propose the perceptual loss, which is defined as the MSE loss over high-level features extracted from the corresponding images. These features, which are extracted using a pretrained 19 layer VGG Network \cite{simonyan2018very}, map raw images to a lower dimensional and representative subspace. Thus, optimizing the perceptual loss better preserves discriminative information and alleviate overfitting. The perceptual loss can be considered as the L2 distance between the feature representations of the generated super-resolved image and ground truth HR image. For \textit{N} training samples we solve:
\vspace{-0.3cm}
\begin{equation}
\begin{split}
L_{per} = \frac{1}{N}\sum\limits_{n=1}^N\frac{1}{C_{j}W_{j}H_{j}}\sum\limits_{c=1}^{C_{j}}\sum\limits_{x=1}^{W_{j}}\sum\limits_{y=1}^{H_{j}}\\(\phi_{j}(I_{n}^{HR})_{c,x,y}-\phi_{j}(G(I_{n}^{LR}))_{c,x,y})^2,
\end{split}
\end{equation}where $\phi_{j}$ stands for feature map of \textit{j}-th convolutional layer and ${C_{j}}$, ${W_{j}}$ and ${H_{j}}$ define the dimensions of the respective feature maps within the VGG19 network.
\subsection {Adversarial Loss}
Since the network cannot learn to recover all high-frequency information by optimizing only the MSE or the perceptual losses, we also add the adversarial loss to the perceptual and the pixel-wise MSE losses to train our proposed network. The adversarial loss described by (1) pushes the solutions move towards the natural image manifold by training the  generator to fool the discriminator by generating photo-realistic images, and training the discriminator to accurately classify "real" images from the generated ones (i.e., fake images). Thus, the estimated solutions reside on the real samples manifold. The adversarial loss $l_{adv}$ defines the probability of the discriminator $D(G(I^{LR}))$ that the reconstructed image $G(I^{LR})$ is a real HR image. Both discriminators as shown in Fig. 2, use the following adversarial loss functions to optimize the network. 
\begin{equation}\begin{split}
L_{adv} &= \min_{G}\max_{D}[E_{I^{HR}\sim P_{train}(I^{HR})}[log D(I^{HR})]\quad \\ & + E_{I^{LR}\sim P_{G}(I^{LR})}[log(1-D(G(I^{LR})))]],\end{split}
\end{equation}where ${P_{train}(I^{HR})}$ and ${P_{G}(I^{LR})}$ define the probability distribution of real high-resolution images and corresponding low-resolution images, respectively.  

\subsection{Detection loss}
YOLOv3 is the combination of three losses: localization, confidence and classification loss.
Equation (6) defines this loss. $1_{ij}^{obj}$ means the object is detected by $j^{th}$ boundary box of grid cell i. $x_{i}$,$y_{i}$, $w_{i}$,$h_{i}$ are the real ground truth bounding box coordinates whereas $\hat{x_{i}}$,$\hat{y_{i}}$, $\hat{w_{i}}$,$\hat{h_{i}}$ are the predicted bounding box coordinates. $C_{i}$ is the box confidence score in cell i, $\hat{C_{i}}$ is the box confidence score for the predicted object:
\begin{equation}\begin{split}
L_{detection} =\lambda_{coord}\sum\limits_{i=0}^{S^2}\sum\limits_{j=0}^B1_{ij}^{obj}(x_i-\hat{x_i})^2+(y_i-\hat{y_i})^2\\+\lambda_{coord}\sum\limits_{i=0}^{S^2}\sum\limits_{j=0}^B1_{ij}^{obj}(\sqrt{w_i}-\sqrt{\hat{w_i}})^2+(\sqrt{h_i}-\sqrt{\hat{h_i}})^2\\+ \sum\limits_{i=0}^{S^2}\sum\limits_{j=0}^B1_{ij}^{obj}l(C_i,\hat{C_i})+\lambda_{noobj}\sum\limits_{i=0}^{S^2}\sum\limits_{j=0}^B1_{ij}^{noobj}l(C_i,\hat{C_i})\\+ \sum\limits_{i=0}^{S^2}1_{i}^{obj}\sum\limits_{c\in classes}l(p_i(c)-\hat{p_i}(c)).\end{split}
\end{equation}
\subsection{Joint Loss Optimization}
Our proposed model can be viewed as a joint learning approach. The network is learning semantic information about targets from the training distribution so that the appearance of the target looks more clear and obvious in super-resolved images to help the detection module. In this section, we show how we combine the detection loss along with the pixel-wise MSE loss, perceptual loss and adversarial loss through an optimization to produce our desired output with full target details. Therefore, to show the dependency of different loss functions, lets assume $W_{SR}$, $W_{VGG}$, $W_{dis}$ and $W_{d}$ denote the parameter set for super-resolution model, pre-trained VGG 19 architecture, discriminator model and detection model, respectively. The parameterized version of the final loss function is as follows:
\begin{equation}
\begin{split}
L = L_{cont}(I_{n}^{LR}; W_{SR}) + \alpha L_{per}(I_{n}^{LR}; W_{SR}, W_{VGG})\\+\beta L_{adv}(I_{n}^{LR}; W_{SR}, W_{dis})\\+\gamma L_{detection}(I_{n}^{LR}; W_{SR}, W_{d}).
\end{split}
\end{equation}
We apply gradient descent algorithm to find the local minimum, and update the network's parameter by calculating the gradient $\nabla{W}$ = $[\nabla{W_{SR}} \nabla{W_{d}}]$ with a learning rate $\eta$.

\subsubsection{Gradient with respect to $W_{d}$}
We calculate $\frac{\partial L}{\partial W_{d}}$ and use the standard 
back propagation algorithm  as the following chain rule holds:

\begin{equation}
\frac{\partial L}{\partial W_{d}} = {\sum_{n=1}^{N}} {\frac{\partial L}{\partial o_{n}}\frac{\partial o_{n}}{\partial W_{d}}},
\end{equation} where $o_{n}$ defines a vector representation of the bounding box coordinates and confidence score.
Again, $\frac{\partial L}{\partial o_{n}}$ involves three terms according to the definition as below:
\begin{equation}
\frac{\partial L}{\partial o_{n}} =  \frac{\partial L_{c}}{\partial o_{n}} + \frac{\partial L_{b}}{\partial o_{n}} + \frac{\partial L_{conf}}{\partial o_{n}},
\end{equation} where $L_{c}$, $ L_{b}$ simply calculate the loss for bounding box coordinates (e.g., center, width and height) and $L_{conf}$ defines bounding box confidence score loss. 
\subsubsection{Gradient with respect to $W_{SR}$}
To update the parameter set for SR model, we consider loss terms associated with the SR reconstruction process and apply gradient descent algorithm to find $\frac{\partial L}{\partial W_{SR}}$. The chain rule holds as follows: 
\begin{equation}
\frac{\partial L}{\partial W_{SR}} =  {\sum_{n=1}^{N}} \frac{\partial L}{\partial G(I_{n}^{LR})}\frac{\partial G(I_{n}^{LR})}{\partial W_{SR}}.
\end{equation} If we set the partial derivative of the loss function with respect to $G(I_{n}^{LR})$ and expand $L$, we get
\begin{equation}\begin{split}
\frac{\partial L}{\partial G(I_{n}^{LR})} =  \frac{\partial L_{cont}}{\partial G(I_{n}^{LR})} + \alpha {\frac{\partial L_{per}}{\partial G(I_{n}^{LR})}} + \beta {\frac{\partial L_{adv} }{\partial G(I_{n}^{LR})}}\\+ \gamma({\frac{\partial L_{c} }{\partial G(I_{n}^{LR})}} + {\frac{\partial L_{b}}{\partial G(I_{n}^{LR})}} + {\frac{\partial L_{conf}}{\partial G(I_{n}^{LR})}}).\end{split}
\end{equation} or, we can also express the above equation as follows:
\begin{equation}\begin{split}
\frac{\partial L}{\partial G(I_{n}^{LR})} =  \frac{\partial L_{cont}}{\partial G(I_{n}^{LR})} + \alpha {\frac{\partial L_{per}}{\partial G(I_{n}^{LR})}}\\+ \beta {\frac{\partial L_{adv} }{\partial G(I_{n}^{LR})}}+ \gamma{\frac{\partial L_{detection}}{\partial G(I_{n}^{LR})}}. \end{split}
\end{equation}
We can summarize the optimization steps in Algorithm 1.

\newcommand\mycommfont[1]{\footnotesize\ttfamily\textcolor{blue}{#1}}
\SetCommentSty{mycommfont}

\SetKwInput{KwInput}{Input}                
\SetKwInput{KwOutput}{Ensure}             
\begin{algorithm}
\DontPrintSemicolon
  
\KwInput{Training samples, I = <$I_{n}^{LR}$, $I_{n}^{HR}$>}
\KwOutput{Model parameters set $W$ = [$W_{SR}, W_{d}$]}
\textbf{while} not converged \textbf{do}
  
t=t+1;

  
  
  
    

calculate the partial derivative $\frac{\partial  L}{\partial W_{d}}$;

calculate the partial derivative $\frac{\partial L}{\partial o_{n}}$;

execute back propagation from top layer to the bottom layer of detection to obtain $\frac{\partial L}{\partial W_{d}}$;

calculate the partial derivative $\frac{\partial L}{\partial G(I_{n}^{LR})}$;

add the $\frac{\partial L_{cont}}{\partial G(I_{n}^{LR})}$, $\frac{\partial L_{per}}{\partial G(I_{n}^{LR})}$ and $\frac{\partial L_{adv}}{\partial G(I_{n}^{LR})}$ to the derivative $\frac{\partial L}{\partial G(I_{n}^{LR})}$ obtained in step 6;

execute back propagation from the last layer to the
first layer of SR to obtain $\frac{\partial L}{\partial W_{SR}}$;

update the parameter $W$ by $W^{t+1}$ = $W^t$ + $\eta  \nabla W$;

\caption{Our proposed Joint-SRVDNet model training}
\end{algorithm}


\section{Training Details}

\subsection{Experimental Data}
We evaluate the performance of our proposed method on three publicly available benchmark datasets: Vehicle Detection in Aerial Imagery (VEDAI) dataset \cite{razakarivony2016vehicle}, xView dataset \cite{lam2018xview} and DOTA dataset \cite{xia2018dota}. In this section, detailed description of the training datasets are provided. Then, we describe the implementation and experimental strategies. 

\subsubsection{Vehicle Detection in Aerial Imagery (VEDAI) Dataset}
The VEDAI dataset is a publicly available benchmark for small 
target recognition especially vehicle detection in aerial images. This dataset has around 1,210 images of two different resolutions such as $1,024\times1,024$ pixels and $512\times512$ pixels. The images mostly contain small vehicles having diverse backgrounds, multiple orientations, lighting/shadowing changes, specularities or occlusions. In addition, it includes nine different classes of vehicles, namely the plane, boat, camping car, car, pick-up, tractor, truck, van, and the other category. We consider all classes as a single class namely 'vehicle' for our task. For training and testing, we split the dataset into 1,100 and 271 images, respectively. The number of samples in our dataset is small for analyzing the proposed network. Therefore, to make the model more robust to different features, we have used different augmentation techniques such as image sharpening and flipping.

\subsubsection{Dataset for Object detection in Aerial images (DOTA)}
DOTA is a large-scale multi-sensor and multi-resolution aerial dataset. This dataset is challenging because of its immense number of object instances from various categories exhibiting a wide variety of scales, orientations and shapes. The dataset contains 2,806 images of varying size ranging from $800\times800$ to $4,000\times4,000$ pixels. We have created patches of size $512\times512$ from the original images. The complex aerial scenes present in this dataset are collected from Google Earth, satellite JL-1 and satellite GF-2. The dataset has fifteen categories of objects namely plane, ship, storage tank, swimming pool, ground track field, harbor, bridge, large vehicle, small vehicle, helicopter, roundabout, soccer ball field, basketball court, baseball diamond and tennis court. We have omitted class swimming pool, ground track field, harbor, bridge, roundabout, soccer ball field, basketball court, baseball diamond and tennis court and unified the remaining six classes as one class 'vehicle'. 

\subsubsection{x-View Dataset}
xView is currently the largest publicly available dataset collected from WorldView-3 satellites. The dataset contains 60 highly imbalanced classes. To overcome the problem of poor detection performance, we have generalized all the classes into one class 'vehicle'. It contains around 1 million objects covering 1,400 $km^{2}$ of the earth surface. The dataset is cropped into smaller patches of $512\times512$. Each pixel corresponds to $0.3\times 0.3$ $m^{2}$ area in the ground. The annotation provided is in geoJSON format and contains information about the bounding boxes for objects present in an image.

\subsection{Implementation Strategies and Training Parameters}
At the beginning, we separately train both sub-networks: super-resolution and detection modules to obtain their network weights which have been used to initialized the joint network of our proposed model. We perform all experiments using 4x upsampling factor
between low- and high-resolution images. To obtain LR images, bicubic kernel is used to downscale the HR images with a scale factor of 4. During implementation, we use input images of size $128\times128$ to super-resolve to $512\times512$. 

To train a deep neural network using a small dataset is troublesome due to the over-fitting problem. One approach to overcome this difficulty is to use data augmentation, specifically sharpening and [horizontal, veritical] flipping.

For the super-resolution network, we adapt the Adam optimizer with a momentum of 0.9 and a batch size of 4. We initially set the learning rate at $10^{-4}$ which decays by a factor of 0.1 after every 5 epochs. For YOLOv3 model, we optimize the network by Adam with a learning rate of $10^{-4}$ and $10^{-6}$ with batch size 16. For non-maximum suppression, the threshold is set to 0.5. Following (6), the  network calculates bounding box loss, coordinate loss, class confidence scores and objectness score for each detection layer. These losses are offset to predict the object probability, class probability and bounding box coordinates for each grid which together represents an object at that grid. Usually the network generates several bounding boxes and selects the bounding box with the highest Intersection over Union (IoU). For each aerial dataset, we train both networks for 10 epochs and achieve satisfactory results. 

For joint-training, we consider the sub-networks together and train it as a unified network. To initialize the overall network, we employ the weights from the independently pre-trained models. We choose Adam as the optimizer by setting initial learning rate as $10^{-4}$. The learning rate decays exponentially with moving average decay of 0.9991. After training for 4 epochs with a mini-batch size 1, we observe
significant improvement in results which verifies that our proposed method has been successfully implemented. We implement the proposed network using tensorflow framework and train it over two NVIDIA Titan XpGPU. Moreover, we explored the effect of varying the hyperparameters ($\alpha$, $\beta$ and $\gamma$) adapted in (7) to further validate the results of our model. The analysis of the hyperparameters has been made on the test dataset and their impact will be discussed in the ablation study. 




\begin{table*}[t]
\renewcommand{\arraystretch}{1.2}
\caption{Comparison of super-resolution architectures for upscale factor 4x  on aerial datasets.}
\centering
\scalebox{0.90}{\begin{tabular}{c|c |c| c|c }
\hline
Dataset & VEDAI-VISIBLE & VEDAI-IR & XVIEW & DOTA\\
\hline
Algorithm &PSNR\hspace{0.35em} MSSIM \hspace{0.4em} UQI  \hspace{0.6em} VIF   &PSNR\hspace{0.35em} MSSIM \hspace{0.4em} UQI  \hspace{0.6em} VIF 
&PSNR\hspace{0.35em} MSSIM \hspace{0.4em} UQI  \hspace{0.6em} VIF   &PSNR\hspace{0.35em} MSSIM \hspace{0.4em} UQI  \hspace{0.6em} VIF\\
\hline
\hline
Bicubic  &\hspace{0.3em}22.060 \hspace{0.6em}0.912 \hspace{0.3em}\hspace{0.3em}\hspace{0.3em}0.945\hspace{0.5em} 0.560  &\hspace{0.3em}22.513 \hspace{0.6em}0.920 \hspace{0.3em}\hspace{0.3em}\hspace{0.3em}0.980\hspace{0.5em} 0.597  &\hspace{0.3em}15.856 \hspace{0.6em}0.419 \hspace{0.3em}\hspace{0.3em}\hspace{0.3em}0.663\hspace{0.5em} 0.416  &\hspace{0.1em}24.617 \hspace{0.6em}0.936 \hspace{0.3em}\hspace{0.3em}\hspace{0.3em}0.963\hspace{0.5em} 0.349\\
SRGAN    &\hspace{0.3em}25.856 \hspace{0.6em}0.918 \hspace{0.3em}\hspace{0.3em}\hspace{0.3em}0.981\hspace{0.5em} 0.607 &\hspace{0.3em}25.876 \hspace{0.6em}0.928 \hspace{0.3em}\hspace{0.3em}\hspace{0.3em}0.988\hspace{0.5em} 0.627  &\hspace{0.3em}17.799 \hspace{0.6em}0.517 \hspace{0.3em}\hspace{0.3em}\hspace{0.3em}0.783\hspace{0.5em} 0.515 &\hspace{0.1em}24.893 \hspace{0.6em}0.941 \hspace{0.3em}\hspace{0.3em}\hspace{0.3em}0.959  \hspace{0.4em}0.514 \\
MsSRGAN  &\hspace{0.3em}26.899 \hspace{0.6em}0.927  \hspace{0.3em}\hspace{0.3em}\hspace{0.3em}0.991\hspace{0.5em} 0.653 &\hspace{0.3em}27.890 \hspace{0.6em}0.939 \hspace{0.3em}\hspace{0.3em}\hspace{0.3em}0.995\hspace{0.5em} 0.683 &\hspace{0.3em}18.838 \hspace{0.6em}0.541 \hspace{0.3em}\hspace{0.3em}\hspace{0.3em}0.794\hspace{0.5em} 0.550  &\hspace{0.1em}28.474 \hspace{0.6em}0.975 \hspace{0.3em}\hspace{0.3em}\hspace{0.3em}0.971\hspace{0.4em} 0.623\\
DenseNet GAN &\hspace{0.3em}29.9\hspace{10.7em} &- &\hspace{0.3em}- &\hspace{0.5em}- \\
\textbf{Joint-SRVDNet (Ours)}     &\hspace{0.3em}\textbf{30.338} \hspace{0.6em}\textbf{0.969}  \hspace{0.3em}\hspace{0.3em}\hspace{0.3em}\textbf{0.995}\hspace{0.5em} \textbf{0.693} &\hspace{0.3em}\textbf{29.227} \hspace{0.6em}\textbf{0.958} \hspace{0.3em}\hspace{0.3em}\hspace{0.3em}\textbf{0.999}\hspace{0.5em} \textbf{0.713}  &\hspace{0.3em}\textbf{20.550} \hspace{0.6em}\textbf{0.617} \hspace{0.3em}\hspace{0.3em}\hspace{0.3em}\textbf{0.795}\hspace{0.5em} \textbf{0.562} &\hspace{0.1em}\textbf{31.360} \hspace{0.6em}\textbf{0.987} \hspace{0.3em}\hspace{0.3em}\hspace{0.3em}\textbf{0.975} \hspace{0.4em}\textbf{0.712} \\
\hline
\end{tabular}
}
\end{table*}

\begin{figure*}[t]
\hspace*{-0.4cm}  
  \centering
      \includegraphics[width=18cm]{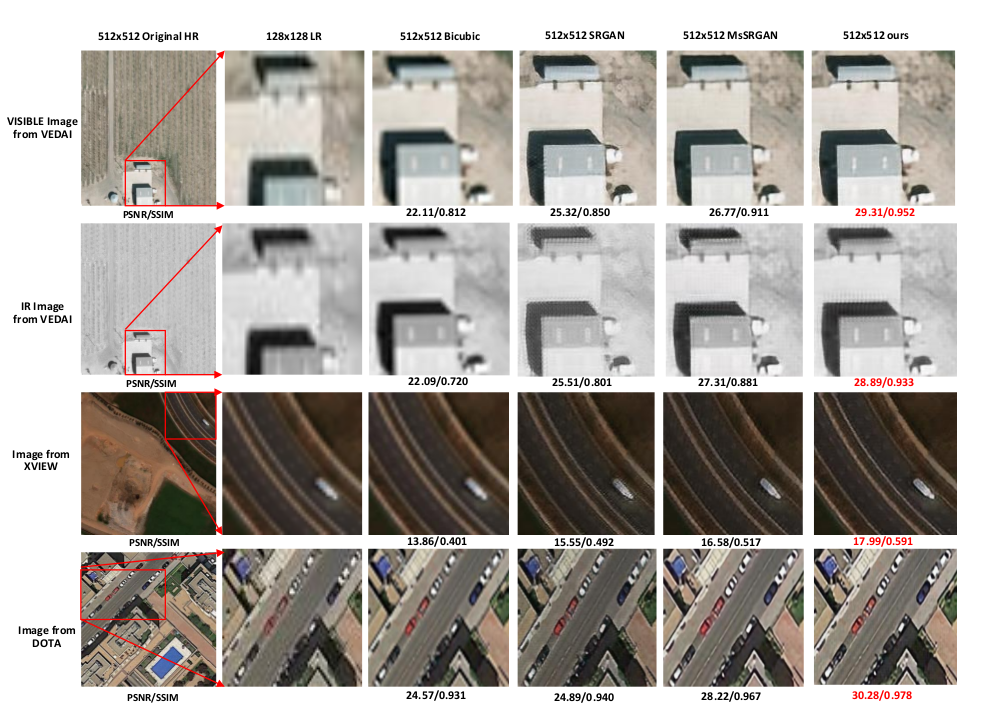}
  \caption{Visual results using Bicubic, SRGAN, MsSRGAN and our proposed model Joint-SRVDNet with scaling factor 4 over VEDAI, xView and DOTA datasets.}
  
\end{figure*}

\section{Experimental Results Analysis}

In this section,  we present
comparative results for both image super-resolution and vehicle detection on several aerial datasets to evaluate the performance of our proposed model. We compare the reconstruction quality of the super-resolved images generated by our proposed network to other methods including bicubic interpolation, SRGAN \cite{ledig2017photo}, MsSRGAN \cite{kazemi2019identity} and DenseNet GAN \cite{bosch2018super} on overhead datasets which were described in the previous section. Then we investigate vehicle detection performance of our network in terms of mean Average Precision (mAP) and F1 score. For more comprehensive performance analysis, we provide precision-recall curve and plot true positive rate (TPR) against false positive rate (FPR).

\begin{table*}[t]
\renewcommand{\arraystretch}{1}

\caption{Comparative detection performance in terms of mean average precision (mAP) and F1-score of the proposed network and existing state-of-the-art approaches. \textcolor{red}{Red} bold indicates
the optimal performance using actual HR imagery and \textcolor{blue}{blue} bold indicates the second optimal performance using SR images generated by our proposed network.}
\centering
\begin{tabular}{c|c |c| c|c }
\hline
Dataset & \hspace{0.1em}VEDAI-VISIBLE & \hspace{0.6em}VEDAI-IR & \hspace{0.6em}XVIEW & \hspace{0.6em}DOTA\\
\hline
Architectures &\hspace{0.1em}mAP@0.5 \hspace{1.2em}F1 score &\hspace{1em}mAP@0.5 \hspace{0.86em}F1 score &\hspace{1em}mAP@0.5 \hspace{0.8em}F1 score &\hspace{1em}mAP@0.5 \hspace{0.8em}F1 score\\
\hline
\hline
Ren, et al. (Z\&F) \cite{ren2015faster} &\hspace{0.6em}32.00              \hspace{2.88em}0.212 &- &-   &-    \\
Girshik, et al. (VGG-16) \cite{girshick2015fast} &\hspace{0.6em}37.30 \hspace{2.89em}0.224 &- &-   &-  \\
Ren, et al. (VGG-16) \cite{ren2015faster} &\hspace{0.6em}40.90 \hspace{2.89em}0.225 &- &-   &-   \\
Zhong, et al. \cite{zhong2017robust} &\hspace{0.6em}50.20 \hspace{2.89em}0.305 &- &-   &-  \\
Chen, et al. \cite{chen2019multiple}&\hspace{0.6em}59.50 \hspace{2.89em}0.451  &- &-   &-  \\
YOLOv3\textunderscore SRGAN\textunderscore 512   &\hspace{0.6em}62.45 \hspace{2.89em}0.591 &\hspace{0.8em}70.10 \hspace{2.89em}0.687  &\hspace{0.8em}53.47 \hspace{2.89em}0.479   &\hspace{1.2em}86.18 \hspace{2.89em}0.837\\
YOLOv3\textunderscore MsSRGAN\textunderscore 512  &\hspace{0.6em}66.74
\hspace{2.89em}0.643 &\hspace{0.8em}74.61 \hspace{2.89em}0.723  &\hspace{0.8em}57.96 \hspace{2.89em}0.494  &\hspace{1.2em}87.02 \hspace{2.89em}0.859\\
YOLOv3\textunderscore SSSDet\textunderscore 512    \cite{mandal2019sssdet}  &45.97\hspace{4.8em}  &- &-             &79.52\hspace{4.3em} \\
Ju, et al. \cite{ju2019simple}  &- &-   &-   &88.63\hspace{4.3em} \\
\textcolor{blue}{\textbf{YOLOv3\textunderscore Joint-SRVDNet\textunderscore 512 (Ours)}}    &\hspace{0.6em}\textcolor{blue}{\textbf{72.46}} \hspace{2.89em}\textcolor{blue}{\textbf{0.702}}&\hspace{0.8em}\textcolor{blue}{\textbf{80.40}} \hspace{2.89em}\textcolor{blue}{\textbf{0.792}} &\hspace{0.8em}\textcolor{blue}{\textbf{61.50}}\hspace{2.89em}\textcolor{blue}{\textbf{0.671}}  &\hspace{1.2em}\textcolor{blue}{\textbf{90.01}} \hspace{2.89em}\textcolor{blue}{\textbf{0.893}}\\
\textcolor{red}{\textbf{YOLOv3\textunderscore HR\textunderscore 512}}  &\hspace{0.6em}\textcolor{red}{\textbf{85.33}} \hspace{2.89em}\textcolor{red}{\textbf{0.826}} &\hspace{0.8em}\textcolor{red}{\textbf{85.66}} \hspace{2.89em}\textcolor{red}{\textbf{0.876}} &\hspace{0.8em}\textcolor{red}{\textbf{66.02}} \hspace{2.89em}\textcolor{red}{\textbf{0.687}} &\hspace{1.2em}\textcolor{red}{\textbf{94.56}} \hspace{2.89em}\textcolor{red}{\textbf{0.933}} \\
\hline
\end{tabular}
\end{table*}

\begin{figure*}%
\hspace*{-0.4cm}   
\centering
\subfloat[]{\includegraphics[width=4.7cm,height=4.5cm]{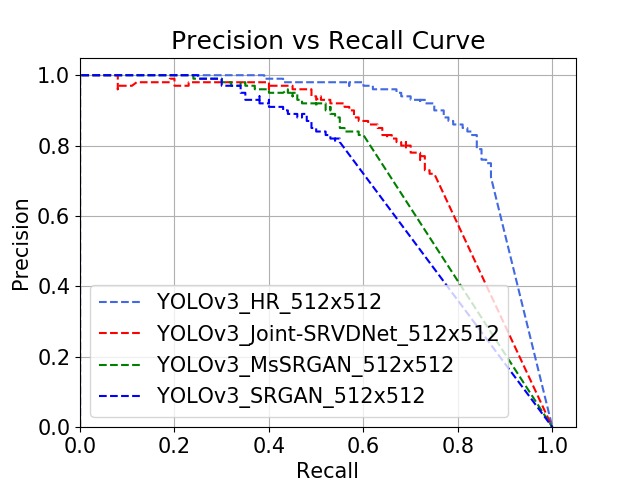}}
\subfloat[]{\includegraphics[width=4.7cm,height=4.5cm]{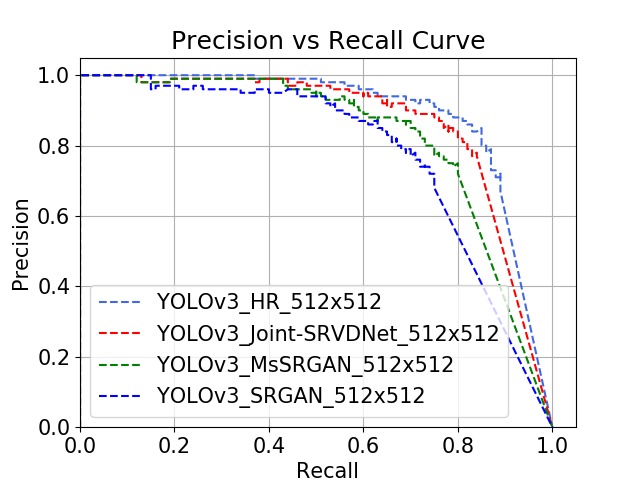}}
\subfloat[]{\includegraphics[width=4.7cm,height=4.5cm]{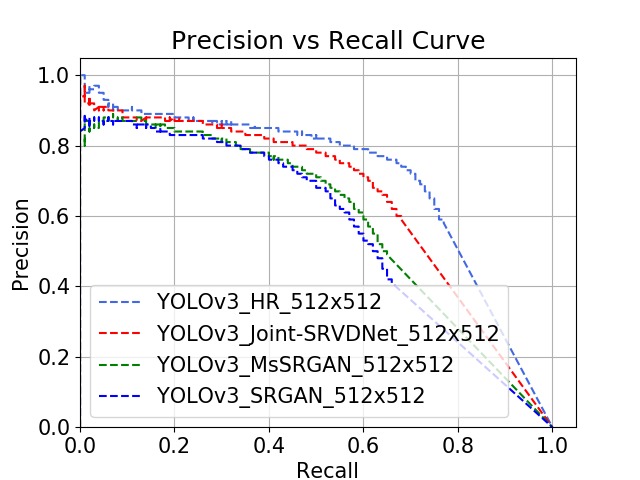}}%
\subfloat[]{\includegraphics[width=4.7cm,height=4.5cm]{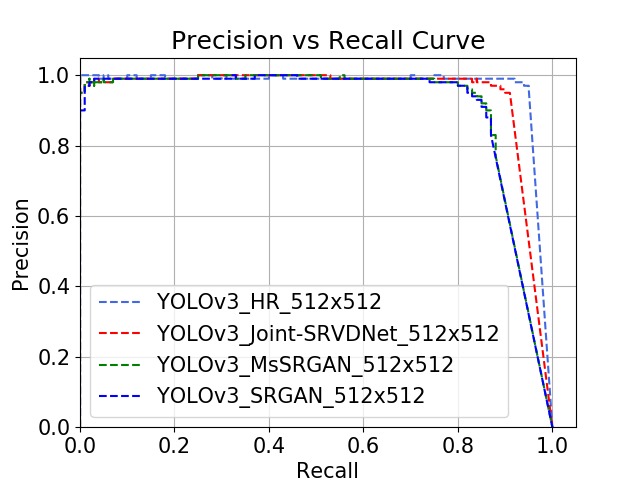}}
\caption{Precision-recall graph of the state-of-the-art object detector YOLOv3 performed on the original 512x512 high-resolution test images and the corresponding super-resolved images generated from SRGAN, MsSRGAN and our proposed Joint-SRVDNet over (a) VEDAI-VISIBLE, (b) VEDAI-IR, (c)xView and (d) DOTA.}%
\end{figure*}

\subsection{Super-Resolution Results}
We have reported the super-resolution results of our experiments using several objective image quality metrics such as Peak Signal-to-Noise ratio (PSNR), Multi-scale Structural  Similarity (MSSIM) \cite{wang2003multiscale}, Universal image Quality Index (UQI) \cite{wang2002universal} and Visual Information Fidelity (VIF) \cite{sheikh2006image} on a validation subset of images for each
dataset. Table \rom{1} shows a comparative analysis of our approach
with other GAN based state-of-the-art techniques.  For comparison, first we include results from  bicubic interpolation method. Then we follow SRGAN architecture; one of the pioneering works on super-resolution using GAN introduced by Ledig et al \cite{ledig2017photo}. As expected, the performance of this network is much better than the previous approaches due to addition of the perceptual loss which enables the network to produce images with sharper edges and features.  After that, we notice, adding multiple intermediatory discriminators to the same generator architecture as the SRGAN helps to generate even higher quality images with more perceptual similarity which often lacks in the generated images from the SRGAN. We refer to this network as MsSRGAN which is actually introduced in \cite{kazemi2019identity} to handle super-resolution for facial images. We utilize this concept and conduct experimients on aerial datasets. We observe slight improvements in the reconstructed SR results and report it for comparison. Moreover, we have also compared our results with DenseNet GAN \cite{bosch2018super} for VEDAI dataset. All these GAN based methods use perceptual loss, MSE loss along with adversarial loss even if they modify their architecture which shows gradual improvement in their solutions. However, they cannot meet the demand of current situation. They are often unable to extract fine texture details of the targets (vehicle) of interest. So, our aim is to produce solutions which contain clear view of our targets with fine-grained details. We design a loss function which incorporates detecton loss along with other losses (perceptual loss, adversarial loss and MSE loss) which helps to reach our goal.  Table \rom{1} shows that our proposed algorithm obtains the highest PSNR, MSSIM, UQI and VIF scores which proves the quantitative effectiveness of our proposed network. 
To show the quality of the super-resolved images specifically for the target regions produced by our network, we select a small area around the targets and show the gradual progression of different SR results  which are visible in Fig. 5.  We have conducted our experiments for 4x enhancement ($128\times128$ to $512\times512$). We can see that in the super-resolved image the selected area around the target and the target itself is getting more close to the original one as bicubic interpolation, SRGAN, MsSRGAN and our network have been applied successively. 
Visual results are showing that recovering high frequency details in low-resolution domain is extremely difficult but it is captured by using our proposed network. 
The ultimate goal of our work is to recover target details which has a great effect on the detection performance.
\subsection{Detection Performance Analysis}
Table \rom{2} summarizes a comparative performance measures of our proposed model and other leading state-of-the-art algorithms in terms of mAP and F1 score for aerial vehicle detection. The mAP values and F1 scores are reported on VEDAI, xView and DOTA datsets for most of the algorithms based on the availability. We calculate the mAP as the average of the maximum precisions at different recall values in the range $(0.0 \sim1.0)$. For each dataset, we show the precision-recall graphs at different IoU thresholds $(0.3 \sim0.7)$ for YOLOv3 performed on super-resolved images generated from SRGAN, MsSRGAN and our proposed network as shown in Fig. 6. We have evaluated all the methods over the same set of test data. we can conclude that our proposed technique is much more stable and robust for aerial vehicle detection in comparison to the current state-of-the-art detection techniques.
\begin{figure*}%
\hspace*{-0.4cm} 
  \centering
    \subfloat[]{{\includegraphics[width=4.7cm,height=4.5cm]{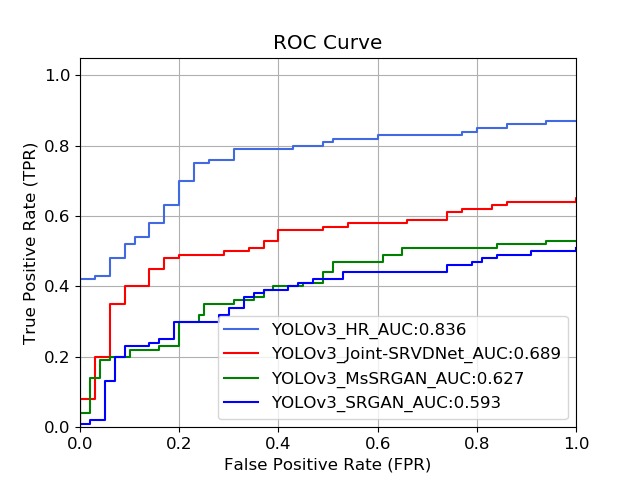}}}%
    \subfloat[]{{\includegraphics[width=4.7cm,height=4.5cm]{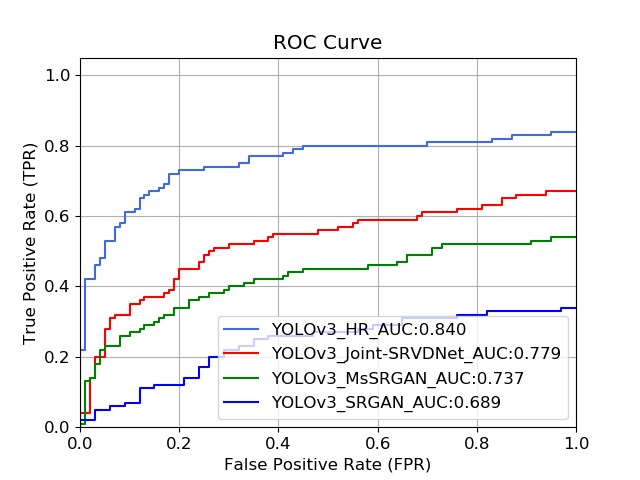}}}
    \subfloat[]{{\includegraphics[width=4.7cm,height=4.5cm]{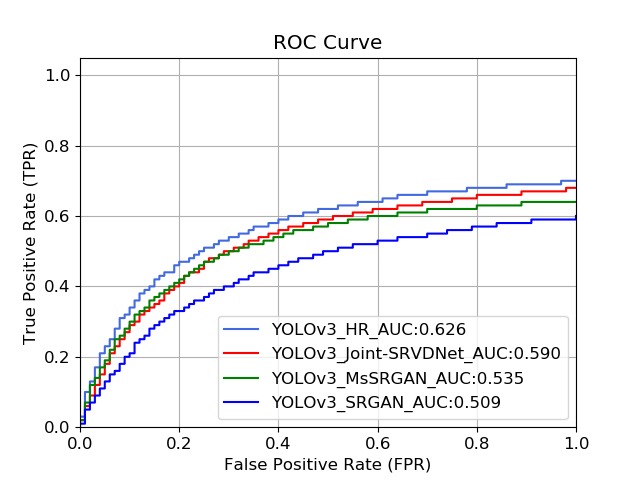}}}%
    \subfloat[]{{\includegraphics[width=4.7cm,height=4.5cm]{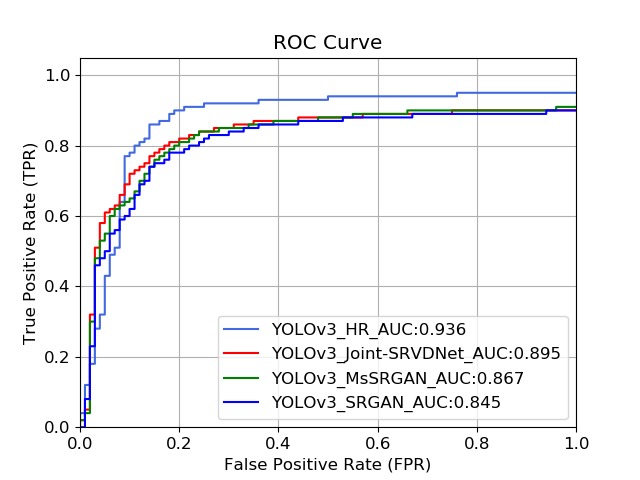}}}
    
     \caption{ROC curves showing true-positive versus false-positive rates of the YOLOv3 object detector performed on the original 512x512 high-resolution test images and the corresponding super-resolved images generated from SRGAN, MsSRGAN and our proposed Joint-SRVDNet over (a) VEDAI-VISIBLE, (b) VEDAI-IR, (c)
xView and (d) DOTA.}%
\end{figure*}
Besides, we include detection results of recent CNN-based detectors: Faster R-CNN \cite{ren2015faster} with Z\&F model, Faster R-CNN \cite{ren2015faster} with VGG-16 model and Fast R-CNN \cite{girshick2015fast} with VGG-16 model for VEDAI dataset. Also, we have compared our detection performance with \cite{zhong2017robust} and most recently proposed detection algorithm \cite{chen2019multiple}.  It is easily noticeable from the results presented in Table \rom{2} that our proposed model demonstrates the best performance compared to these detection methods and yields the 2nd best  mAP (72.46\%) and F1-Score (0.702) for VEDAI. 
For comparison with the current DCNN based approaches, we include the results of SSSDet \cite{mandal2019sssdet} reported in their publications for VEDAI and DOTA as they claim to achieve the most competitive results on such datasets.  We observe that detection performance of our method on VEDAI and DOTA datasets is extremely good compared to \cite{mandal2019sssdet} in terms of mAP. As shown in Table \rom{2}, the performance of our proposed scheme is 26.49\%  and 10.49\% higher than \cite{mandal2019sssdet} for VEDAI-VISIBLE and DOTA datasets respectively.

\begin{table*}[t]
\renewcommand{\arraystretch}{1}
\caption{Super-resolution results of our proposed model using different hyperparameter settings for upscale factor 4x  on the aerial test datasets. \textcolor{magenta}{Magenta} bold indicates the  optimal SR results generated by our proposed network.}
\centering
\scalebox{0.89}{\begin{tabular}{c|c |c| c|c }
\hline
Dataset & VEDAI-VISIBLE & VEDAI-IR & XVIEW & DOTA\\
\hline
Hyperparameter Settings &PSNR\hspace{0.35em} MSSIM \hspace{0.4em} UQI  \hspace{0.6em} VIF   &PSNR\hspace{0.35em} MSSIM \hspace{0.4em} UQI  \hspace{0.6em} VIF 
&PSNR\hspace{0.35em} MSSIM \hspace{0.4em} UQI  \hspace{0.6em} VIF   &PSNR\hspace{0.35em} MSSIM \hspace{0.4em} UQI  \hspace{0.6em} VIF\\
\hline
\hline
$\alpha =  2\times 10^{-6}$, $\beta = 10^{-2}$, $\gamma = 10^{-2}$  \hspace{0.1em} &\hspace{0.3em}27.060 \hspace{0.6em}0.812 \hspace{0.3em}\hspace{0.3em}\hspace{0.3em}0.745\hspace{0.5em} 0.690  &\hspace{0.3em}26.513 \hspace{0.6em}0.720 \hspace{0.3em}\hspace{0.3em}\hspace{0.3em}0.780\hspace{0.5em} 0.697  &\hspace{0.3em}17.856 \hspace{0.6em}0.529 \hspace{0.3em}\hspace{0.3em}\hspace{0.3em}0.523\hspace{0.5em} 0.436  &\hspace{0.1em}24.327 \hspace{0.6em}0.845 \hspace{0.3em}\hspace{0.3em}\hspace{0.3em}0.813\hspace{0.5em} 0.457\\
\scalebox{0.90}{\textcolor{magenta}{$\boldsymbol{\alpha
= 2\times 10^{-6}}$
, $\boldsymbol{\beta = 10^{-3}}$, $\boldsymbol{\gamma = 10^{-3}}$}}\hspace{0.1em} 
&\hspace{0.3em}\textcolor{magenta}{\textbf{30.338}} \hspace{0.6em}\textcolor{magenta}{\textbf{0.969}} \hspace{0.3em}\hspace{0.3em}\hspace{0.3em}\textcolor{magenta}{\textbf{0.995}}\hspace{0.5em} \textcolor{magenta}{\textbf{0.693}} &\hspace{0.3em}\textcolor{magenta}{\textbf{29.227}} \hspace{0.6em}\textcolor{magenta}{\textbf{0.958}} \hspace{0.3em}\hspace{0.3em}\hspace{0.3em}\textcolor{magenta}{\textbf{0.999}}\hspace{0.5em} \textcolor{magenta}{\textbf{0.713}} &\hspace{0.3em}\textcolor{magenta}{\textbf{20.550}} \hspace{0.6em}\textcolor{magenta}{\textbf{0.617}} \hspace{0.3em}\hspace{0.3em}\hspace{0.3em}\textcolor{magenta}{\textbf{0.795}}\hspace{0.5em} \textcolor{magenta}{\textbf{0.562}} &\hspace{0.1em}\textcolor{magenta}{\textbf{31.360}} \hspace{0.6em}\textcolor{magenta}{\textbf{0.987}} \hspace{0.3em}\hspace{0.3em}\hspace{0.3em}\textcolor{magenta}{\textbf{0.975}} \hspace{0.4em}\textcolor{magenta}{\textbf{0.712}} \\
$\alpha = 2\times 10^{-6}$ , $\beta = 10^{-2}$, $\gamma = 10^{-4}$
\hspace{0.1em} &\hspace{0.3em}26.746 \hspace{0.6em}0.723 \hspace{0.3em}\hspace{0.3em}\hspace{0.3em}0.716\hspace{0.5em} 0.705 &\hspace{0.3em}25.976 \hspace{0.6em}0.723 \hspace{0.3em}\hspace{0.3em}\hspace{0.3em}0.789\hspace{0.5em} 0.778  &\hspace{0.3em}16.799 \hspace{0.6em}0.427 \hspace{0.3em}\hspace{0.3em}\hspace{0.3em}0.654\hspace{0.5em} 0.515 &\hspace{0.1em}24.212 \hspace{0.6em}0.841 \hspace{0.3em}\hspace{0.3em}\hspace{0.3em}0.849  \hspace{0.4em}0.524 \\
\hline
\end{tabular}}
\end{table*}

\begin{table*}[t]
\renewcommand{\arraystretch}{1}
\caption{Vehicle detection results in terms of mean average precision (mAP) and F1-score of our proposed model using different hyperparameter settings on the aerial test datasets.   \textcolor{cyan}{Cyan} bold indicates the second optimal performance using SR images generated by our proposed network.}
\centering
\begin{tabular}{c|c |c| c|c }
\hline
Dataset & \hspace{0.1em}VEDAI-VISIBLE & \hspace{0.6em}VEDAI-IR & \hspace{0.6em}XVIEW & \hspace{0.6em}DOTA\\
\hline
Hyperparameters Settings &\hspace{0.1em}mAP@0.5 \hspace{1.2em}F1 score &\hspace{1em}mAP@0.5 \hspace{0.86em}F1 score &\hspace{1em}mAP@0.5 \hspace{0.8em}F1 score &\hspace{1em}mAP@0.5 \hspace{0.8em}F1 score\\
\hline
\hline
$\alpha = 2\times 10^{-6}$, $\beta = 10^{-2}$, $\gamma = 10^{-2}$ \hspace{0.1em} &\hspace{0.6em}68.89 \hspace{2.88em}0.678 &\hspace{0.6em}77.78 \hspace{2.88em}0.756 &\hspace{0.6em}59.61 \hspace{2.88em}0.556   &\hspace{0.6em}88.59 \hspace{2.88em}0.778  \\
\scalebox{0.90}{\textcolor{cyan}{$\boldsymbol{\alpha = 2\times 10^{-6}}$, $\boldsymbol{\beta = 10^{-3}}$, $\boldsymbol{\gamma = 10^{-3}}$}} \hspace{0.1em} &\hspace{0.6em}\textcolor{cyan}{\textbf{72.46}} \hspace{2.89em}\textcolor{cyan}{\textbf{0.702}}
&\hspace{0.8em}\textcolor{cyan}{\textbf{80.40}} \hspace{2.89em}\textcolor{cyan}{\textbf{0.792}} &\hspace{0.8em}\textcolor{cyan}{\textbf{61.50}}
\hspace{2.89em}\textcolor{cyan}{\textbf{0.671}}  &\hspace{0.8em}\textcolor{cyan}{\textbf{90.01}} \hspace{2.88em}\textcolor{cyan}{\textbf{0.893}}
\\
$\alpha = 2\times 10^{-6}$, $\beta = 10^{-2}$, $\gamma = 10^{-4}$  \hspace{0.1em} &\hspace{0.6em}69.90 \hspace{2.89em}0.685 &\hspace{0.6em}78.79 \hspace{2.89em}0.771 &\hspace{0.6em}58.88 \hspace{2.89em}0.521   &\hspace{0.6em}89.12 \hspace{2.89em}0.789 \\
\hline
\end{tabular}
\end{table*}

Again, compared with the detection performance of super-resolved images generated from the existing most resent MsSRGAN based SR architecture,  our method has achieved almost 5.75\% higher mAP and 7\% better F1 score for both VEDAI-VISIBLE and VEDAI-IR images. Moreover, for both dataset, we observe that the detection performance of our network (indicated by blue bold in Table \rom{2}) is also close to the optimal performance of the detector using original HR imagery, which is shown at the bottom row of Table \rom{2}. We also report mAP and F1 score for the xView satellite images which is very challenging as it contains extremely small targets in the image. Due to the low-resolution, targets do not contain detailed information which might help the detection task. As a result we cannot achieve satisfactory performance like other two datasets. However, still we have achieved 3.54\% higher mAP  and 2\% better F1 score than the performance of super-resolved images from  MsSRGAN and it is also close to the detection performance of the original 512x512 high resolution images. We also investigate our model's performance on DOTA dataset. During experiments, we notice a great improvement in detection performance for this dataset as shown in Fig. 6(d) and fourth column of Table \rom{2}. The targets in this dataset seem to have the best appearance quality among two other datasets which has contributed to secure high detection performance. Therefore, we obtain promising results compared to \cite{koester2019comparison} as well as for all the other algorithms. 

In addition, Fig. 6 helps to analysis the relationship between precision and recall rate for all datasets. It is obvious from the precision-recall plots that, our method (YOLOv3\textunderscore Joint-SRVDNet\textunderscore 512x512 in red curve) is significantly better than the other GAN based methods (YOLOv3\textunderscore MsSRGAN\textunderscore 512x512 in green curve and YOLOv3\textunderscore SRGAN\textunderscore 512x512 in blue curve) and more specifically, the performance gain is comparable to the detection performance of the original 512x512 high resolution images.

However, some important information might be missing if we only depend on precision-recall metric and F1 scores to determine the performance of our proposed method. For more robust analysis, we focus on plotting receiver operating characteristic curve (ROC) to study the characteristics of detection results. ROC curve can be drawn by plotting TPR against FPR at different thresholds. ROC curve reflects the relationship between TPR and FPR which may help to compare our method to other detection approaches. To show the comparative detection results for all datasets similar to Fig. 6, we have plotted ROC curves  for different detection methods in Fig. 7.
Furthermore, we calculate the area under the ROC curve known as AUC which can be considered as another important metric to evaluate detection accuracy. According to the analysis of detection results of different frameworks in terms of AUC, the performance of our proposed system is 6.2\%, 4.2\%, 5.5\% and 2.8\% higher in comparison to detection performance of super-resolved images generated from MsSRGAN over VEDAI-VISIBLE, VEDAI-IR, xView and DOTA dataset, respectively.

\section{Ablation Study}
To achieve the best version of our proposed model, we made several experiments through changing the value of hyperparameters to see the impact of the hyperparameter changes on the original version of our work. We have summarized the analysis in Table \rom{3} and \rom{4}.

\subsection{Hyperparameter analysis} 
We analyze the values of $\alpha$, $\beta$ and $\gamma$ adapted in (7) in order to obtain better quantitative results in aerial datasets. In (7), we have used  $\alpha$, $\beta$ and $\gamma$ as weight factors to numerically balance the magnitude of different losses which accelerates the total loss convergence. The network can benefit from the relative influence of different loss functions, which is somehow guided by the weight factors. Since there is no rule of choosing the optimum parameters for the model, we conduct a series of experiments to find out the optimal parameters of the proposed model. We observe that the optimal values lead the training to generate real-looking images with full target details (edges, sharpness, perceptual features, etc.), that has been already reported in the experimental result analysis section. In Table \rom{3} and \rom{4}, we show the average accuracy of our model varying these hyperparameters on several aerial test datasets. 

Among the above settings, we report the results for the second setting (indicated by bold \textcolor{magenta}{Magenta}, \textcolor{cyan}{Cyan}) in Table 1, Table 2, and Fig 5, Fig 6, and Fig 7 as it provides the best results that is almost comparable to the original HR.

\section{Conclusion}
To address the challenge of detecting small targets (vehicles) in aerial images, we propose an approach
that jointly optimizes super-resolution and detection modules. The purpose of our algorithm is to generate high quality super-resolved images from lower-resolution images, so that larger areas can be surveilled with minimal degradation in detection performance. With extensive experiments we demonstrated that our proposed joint network is able to learn and extract features from low-resolution domain which reflects in the generated super-resolved images produced by the network and helps to improve detection performance. Most importantly, the proposed network has two vital contributions: for super-resolution task, using multi-scale GAN approach instead of classical SRGAN approach makes the detection task easier by adding more details in the super-resolved images which is essential to locate objects in the aerial images. Second, network's total loss integrates detection loss during super-resolution training which helps the SR module to specially learn the target area so that those specific area gets more obvious in the final super-resolution results. To evaluate our model's performance we conduct experiments on several publicly available datasets and the results indicate that compared with the leading state-of-the-art super-resolution and detection approaches, our proposed network achieves impressive results and it may have great impact on remote sensing community.

{
\bibliographystyle{IEEEtran}
\bibliography{FINAL_VERSION.bib}
}

\begin{IEEEbiography}[{\includegraphics[width=1in,height=10in,clip,keepaspectratio]{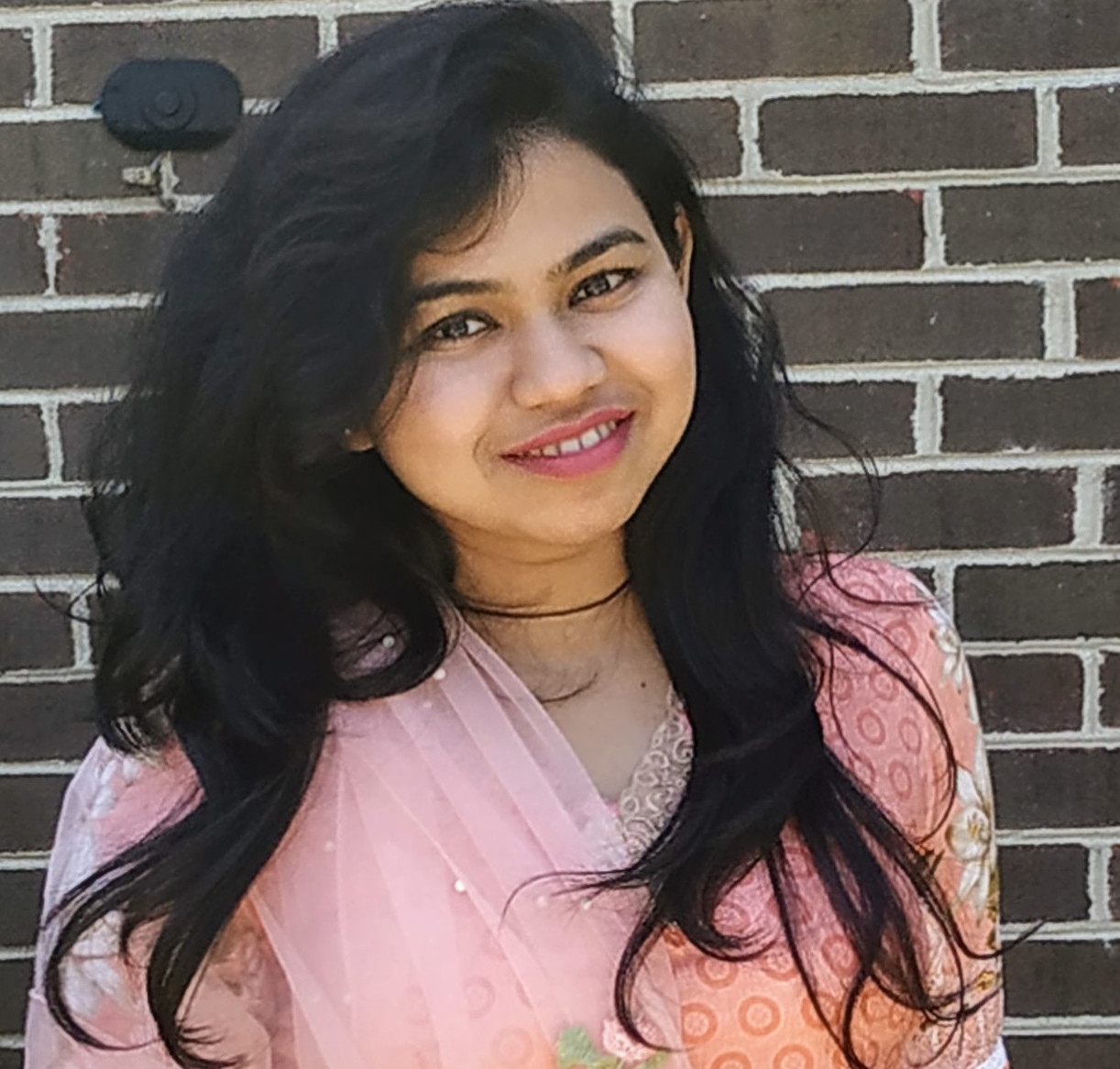}}]{Moktari Mostofa} was born in Dhaka, Bangladesh in 1993. She received the B.Sc degree in Electrical and Electronic Engineering and M.Sc degree in Communication and Signal Processing from the University of Dhaka, Bangladesh in 2016 and 2018 respectively. Currently, she is doing Ph.D. in Electrical Engineering at West Virginia University, WV, USA. 

Since 2018, she has been an research assistant with the Deep Learning Lab and doing research with Professor Nasser M. Nasrabadi. Her research includes mostly the applications of deep learning, machine learning and image processing.   
\end{IEEEbiography}
\begin{IEEEbiography}[{\includegraphics[width=1in,height=1.25in,clip,keepaspectratio]{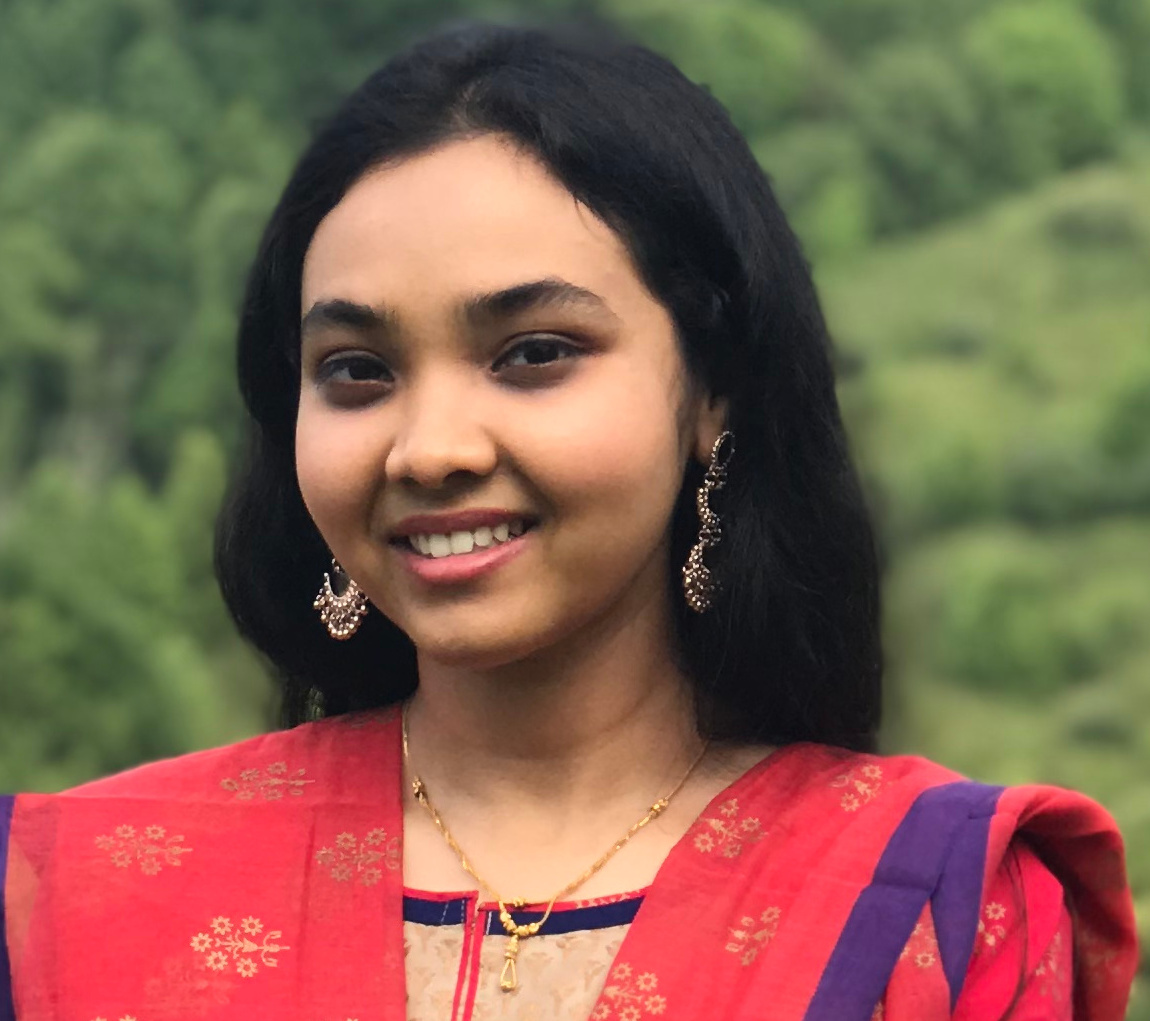}}]{Syeda Nyma Ferdous}  received her B.S. degree in Computer Science from Military Institute of Science and Technology, Bangladesh, in 2013. She completed her M.Sc in Computer Science from Bangladesh University of Engineering and Technology, Bangladesh, in 2017.  She is currently pursuing a Ph.D in the Lane Department of Computer Science  and Electrical Engineering, West Virginia University (WVU), USA. Her research interests include deep learning, computer vision, image processing and machine learning
\end{IEEEbiography}
\begin{IEEEbiography}[{\includegraphics[width=1.01in,height=2.5in,clip,keepaspectratio]{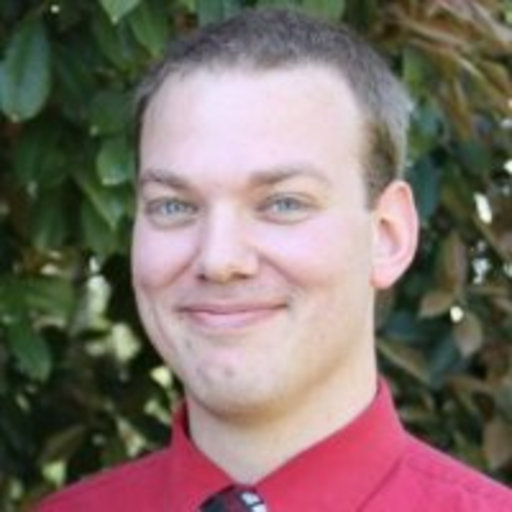}}]{Benjamin S. Riggan}(S'12--M'14) received the B.S. degree in computer engineering from North Carolina State University, in 2009, and the M.S. and Ph.D. degrees in electrical engineering from North Carolina State University, in 2011 and 2014, respectively.  After finishing his Ph.D., he was awarded a postdoctoral fellowship at the U.S. Army Research Laboratory's Image Processing Branch, where he worked on face recognition. He was a Post-Doctoral Fellow with the U.S. Army Research Laboratory. He worked for the Networked Sensing and Fusion Branch of the U.S. Army Research Laboratory. 
His interests are in areas of biometrics and fusion, which leverage his expertise in image/signal processing, computer vision, and machine learning. He has published many papers, and a book, spanning the subjects of handwriting recognition, face recognition, and fusion. Currently, he is an assistant professor with Department of Electrical and Computer Engineering, University of Nebraska-Lincoln, Lincoln, NE, USA. His current research involves cross-spectrum face recognition and spatial-temporal data fusion for target detection/recognition. 
\end{IEEEbiography}
\begin{IEEEbiography}[{\includegraphics[width=1in,height=1.25in,clip,keepaspectratio]{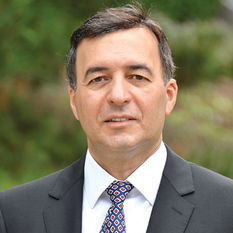}}]{Nasser M. Nasrabadi} (S'80--M'84--SM'92--F'01) received the B.Sc. (Eng.) and Ph.D. degrees in Electrical Engineering from the Imperial College of Science and Technology, University of London, London, U.K., in 1980 and 1984, respectively. In 1984, he was with IBM, U.K., as a Senior Programmer. From 1985 to 1986, he was with the Philips Research Laboratory, New York, NY, USA, as a member of the Technical Staff. From 1986 to 1991, he was an Assistant Professor with the Department of Electrical Engineering, Worcester Polytechnic Institute, Worcester, MA, USA. From 1991 to 1996, he was an Associate Professor with the Department of Electrical and Computer Engineering, State University of New York at Buffalo, Buffalo, NY, USA. From 1996 to 2015, he was a Senior Research Scientist with the U.S. Army Research Laboratory. Since 2015, he has been a Professor with the Lane Department of Computer Science and Electrical Engineering. His current research interests include image processing, computer vision, biometrics, statistical machine learning theory, sparsity, robotics, neural networks and image processing. He is a fellow of the international society for optics and photonics, ARL and SPIE. He has served as an Associate Editor for the IEEE Transactions on Image Processing, the IEEE Transactions on Circuits, Systems and Video Technology, and the IEEE Transactions on Neural Networks.
\end{IEEEbiography}
\EOD
\end{document}